\documentclass[letterpaper, 10 pt, journal, twoside]{IEEEtran}
\ifCLASSINFOpdf
\else
\fi

\usepackage{amsmath} 
\usepackage{amssymb}  
\usepackage{graphicx}
\usepackage{commath}
\usepackage{makecell}
\usepackage{xcolor}
\usepackage{hyperref}
\usepackage{multirow}

\DeclareMathOperator{\Tr}{Tr}
\newcommand{\redbold}[1]{{\color{blue} \textbf{#1}}}

\begin{document}
%
\title{Homography-Based Loss Function for Camera Pose Regression}
%
%
%

\author{Clémentin Boittiaux$^{1,2,3}$, Ricard Marxer$^{2}$, Claire Dune$^{3}$, Aurélien Arnaubec$^{1}$ and Vincent Hugel$^{3}$
\thanks{Manuscript received: November 23, 2021; Revised March 9, 2022; Accepted April 3, 2022.}
\thanks{This paper was recommended for publication by Editor Sven Behnke upon evaluation of the Associate Editor and Reviewers' comments.
This work was supported by Ifremer} 
\thanks{$^{1}$Ifremer, Zone Portuaire de Bregaillon, La Seyne-sur-Mer, France
        {\tt\small firstname.name@ifremer.fr}}%
\thanks{$^{2}$Université de Toulon, Aix Marseille Univ, CNRS, LIS, Toulon, France}%
\thanks{$^{3}$Université de Toulon, COSMER, Toulon, France}%
\thanks{Digital Object Identifier (DOI): see top of this page.}
}
%
%

\markboth{IEEE Robotics and Automation Letters. Preprint Version. Accepted April, 2022}
{Boittiaux \MakeLowercase{\textit{et al.}}: Homography-based loss function} 

%



\maketitle

\begin{abstract}
Some recent visual-based relocalization algorithms rely on deep learning methods to perform camera pose regression from image data. This paper focuses on the loss functions that embed the error between two poses to perform deep learning based camera pose regression. Existing loss functions are either difficult-to-tune multi-objective functions or present unstable reprojection errors that rely on ground truth 3D scene points and require a two-step training. To deal with these issues, we introduce a novel loss function which is based on a multiplane homography integration. This new function does not require prior initialization and only depends on physically interpretable hyperparameters. Furthermore, the experiments carried out on well established relocalization datasets show that it minimizes best the mean square reprojection error during training when compared with existing loss functions.
\end{abstract}


%
\IEEEpeerreviewmaketitle

\section{Introduction}
%
%
%
%
\IEEEPARstart{T}{his} paper addresses the problem of relocating a robot in a place that it has previously visited. In many cases, GPS localization is either not available (\textit{e.g.}, in an underwater environment), very noisy (\textit{e.g.}, cities with high buildings), or insufficient for the target application (\textit{e.g.}, relocalizing a robot inside a building). Under these constraints, it may be possible to estimate a more accurate position for the robot by exploiting its visual observations. The problem, termed visual-based localization~\cite{piasco_survey_2018}, consists in retrieving the pose of a camera in a known 3D scene from an instantaneous image. This problem also appears in loop closure when performing Simultaneous Localization And Mapping (SLAM)~\cite{piasco_survey_2018}.

Until the 2010s, this was solved with classic computer vision methods leveraging engineered descriptors like SIFT or ORB, and PnP/RANSAC schemes~\cite{mur-artal_orb-slam_2015, sattler_efficient_2017}. With recent advances in machine learning, some steps of these methods were greatly outperformed by deep learning based approaches. For instance, learned features have proven to be much more robust and accurate than the aforementioned descriptors for tasks like image classification and image retrieval~\cite{szegedy_going_2014, sandler_mobilenetv2_2019}. By bootstrapping such features, end-to-end gradient-based pose regressors emerged~\cite{kendall_posenet_2016, kendall_geometric_2017, brachmann_dsac_2018, brachmann_learning_2018}. When quantifying the error from a reference pose, these techniques are all confronted with the need to embed the difference between two poses in $SE(3)$ into a scalar. Some more recent approaches focus on learning robust features thanks to deep learning based algorithms and solve the pose estimation with classical computer vision methods~\cite{sarlin_coarse_2019, sarlin_back_2021}.

\begin{figure}[t]
    \centering
    \includegraphics[width=\columnwidth]{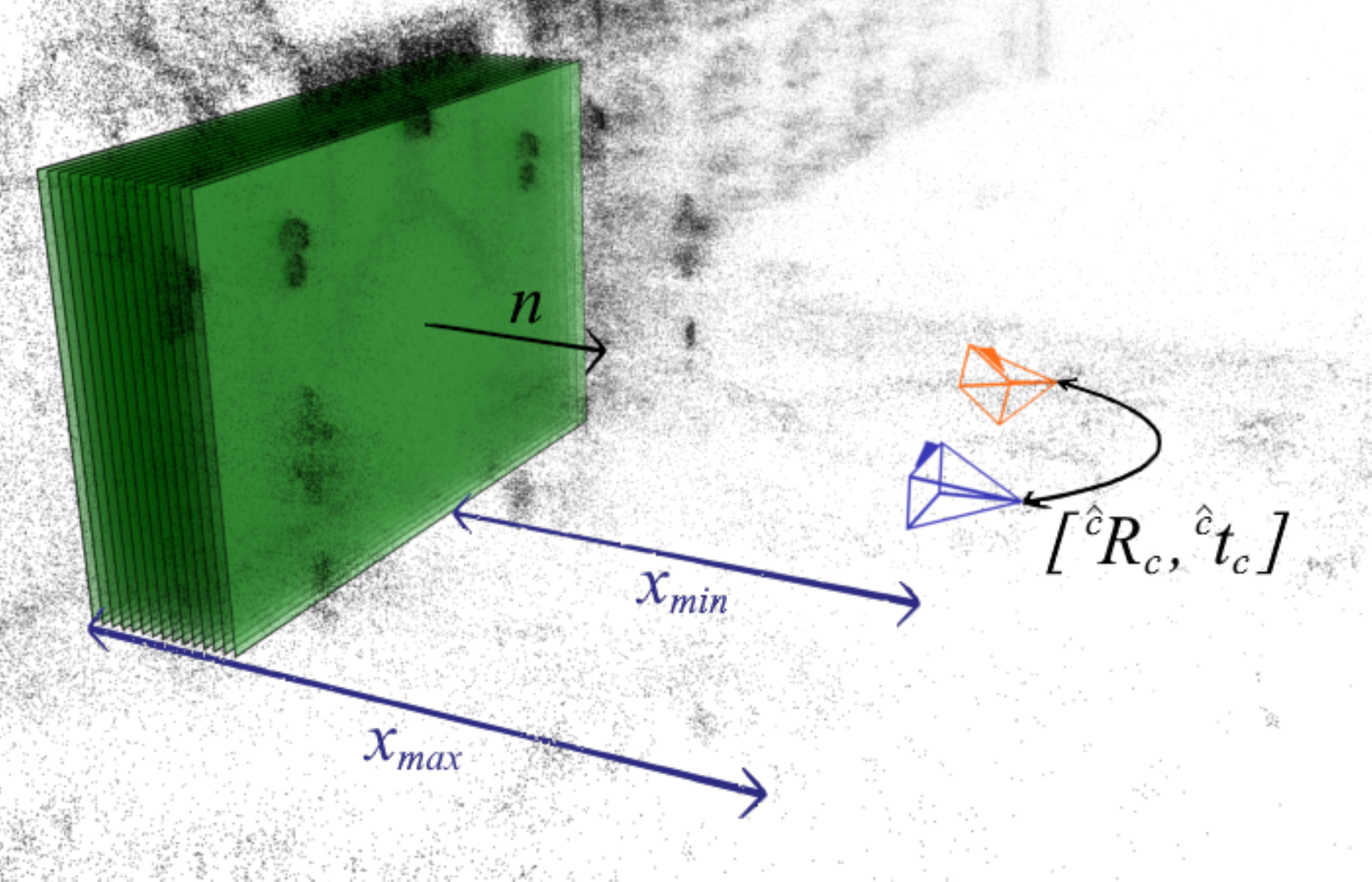}
    \caption{Set of parallel virtual planes (green) between $x_{min}$ and $x_{max}$ depth used to compute our homography loss that quantifies the pose error $[^{\hat{c}}\pmb{R}_c,\, ^{\hat{c}}\pmb{t}_c]$ between ground truth (blue) and estimated (orange) cameras. The planes' normal $\pmb{n}$ and the ground truth camera's optical axis are co-linear. Planes are infinite, but for the sake of visualization they are represented as rectangles.}
    \label{fig:homography_loss}
\end{figure}

\begin{figure*}[t]
    \centering
    \begin{tabular}{ccc}
    \includegraphics[width=0.3\linewidth]{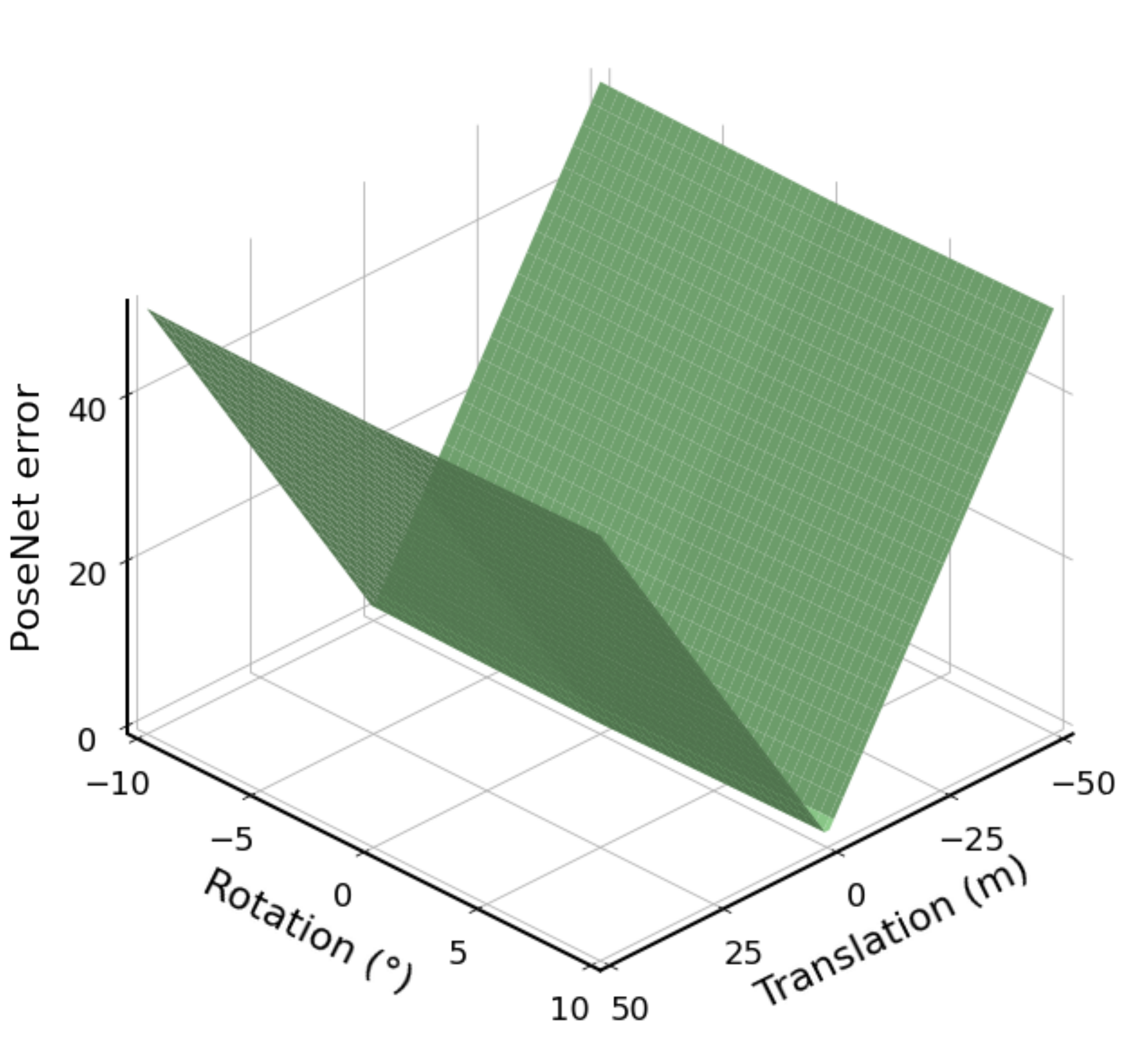}&
    \includegraphics[width=0.3\linewidth]{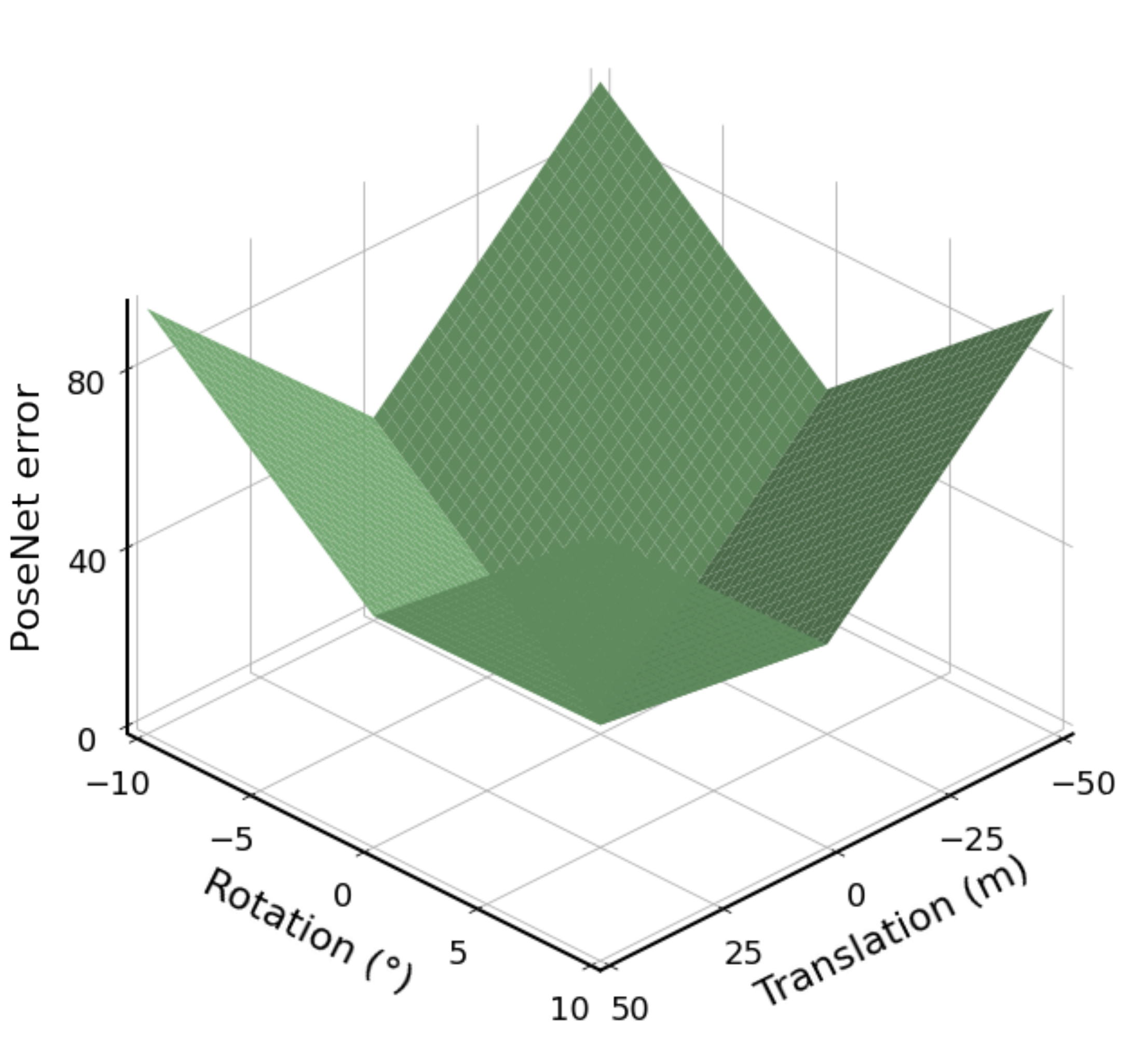}&
    \includegraphics[width=0.3\linewidth]{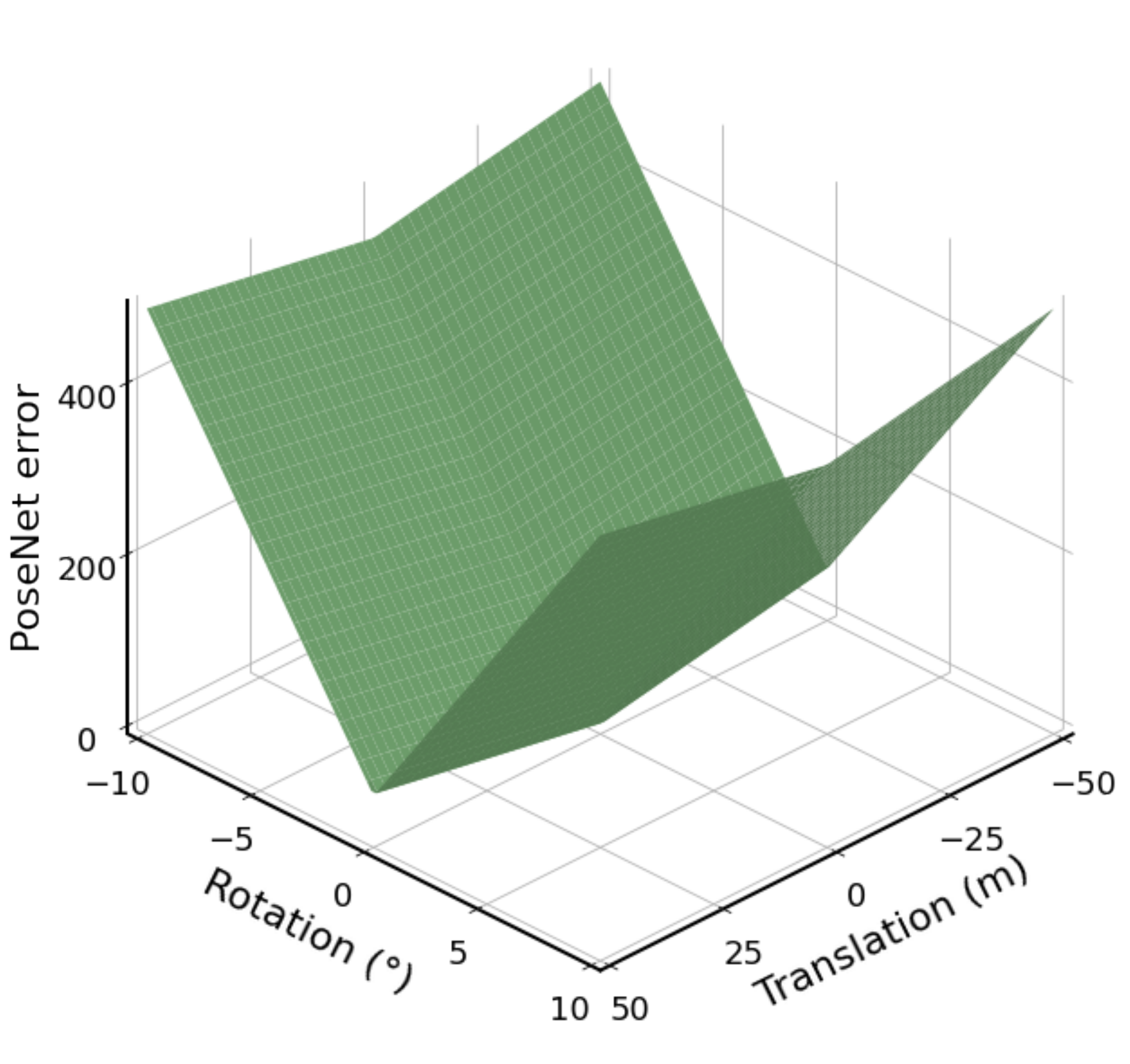}\\
    (a) $\beta=5$&(b) $\beta=500$&(c) $\beta=5000$
    \end{tabular}
    \caption{Effect of $\beta$ on the PoseNet loss (\ref{eq:posenetloss}) for a particular scene: (a) low $\beta$ values lead to high variation of error with respect to translation and little change with respect to rotation; (b) well-chosen $\beta$ leads to a clear local minimum around the optimal parameters; (c) a large $\beta$ induces a small variation of translation.}
    \label{fig:posenet_3d}
\end{figure*}

This paper focuses on the loss functions used in deep learning for camera pose regression. One of the main challenges in defining an error in $SE(3)$ is to properly weight the translation and rotation components in the final error. Some losses rely on a scene-agnostic linear combination of these components, that is, they weight translation and rotation errors regardless of what the camera observes~\cite{kendall_posenet_2016, brachmann_dsac_2018, kendall_multi-task_2018}. Their physical meaning is often difficult to understand because rotation and translation do not lie on the same group. Moreover, setting the explicit weight for each component usually requires empirical fine-tuning and depends on the scene~\cite{kendall_posenet_2016, brachmann_dsac_2018}. Other losses implicitly weight the two components by minimizing the distance between the reprojection of 3D scene points onto the cameras' image planes. These approaches face some issues regarding their initialization and their differentiability in $SE(3)$~\cite{kendall_geometric_2017}. To deal with the problems described above, this paper introduces a new loss function that requires no prior initialization and depends on intuitive parameters. This new loss function approximates the reprojection error as poses get close. We define planes parallel to the ground truth reference image plane for a given range of distances covering the scene. We then express the error in terms of the homographies between the ground truth and estimated poses induced by these planes. The experiments carried out on popular relocalization datasets show that this new loss minimizes best the reprojection error during training when compared with previously mentioned losses.

Section \ref{sec:litterature} reviews existing loss functions~\cite{kendall_posenet_2016, kendall_geometric_2017, brachmann_dsac_2018, kendall_multi-task_2018} and discusses their theoretical and practical characteristics. Section \ref{sec:method} describes the new loss function for camera pose regression. Section \ref{sec:expe} details the implementation of the new loss and the existing losses mentioned in Section \ref{sec:litterature} in a simple end-to-end camera relocalization pipeline using a single backbone architecture\footnote{\raggedright Our code will be made publicly available at \href{run:https://github.com/clementinboittiaux/homography-loss-function}{github.com/clementinboittiaux/homography-loss-function}}. Section \ref{sec:discussion} is dedicated to the performance evaluation of all the losses on the Cambridge~\cite{kendall_posenet_2016} and 7-Scenes~\cite{shotton2013_7scenes} datasets, and to the discussion of the results.

%

\section{Existing pose regression loss functions}
\label{sec:litterature}

Previous computer vision approaches to camera pose regression have focused on active search methods~\cite{sattler_efficient_2017}. They are purely geometry-based and rely on image keypoints extraction. With the recent renaissance of neural networks and deep learning techniques, we have seen a paradigm shift arise: end-to-end pose regression with convolutional neural networks allow to directly infer a pose from training image data~\cite{kendall_posenet_2016, kendall_geometric_2017, kendall_multi-task_2018}. These methods are often far more robust to noise and easy to use, but less accurate~\cite{piasco_survey_2018}. Nowadays, state-of-the-art algorithms try to keep the best of both approaches by replacing some steps of geometry-based methods using deep learning techniques~\cite{brachmann_dsac_2018, brachmann_learning_2018, sarlin_coarse_2019, sarlin_back_2021}.

Visual-based relocalization aims at finding the pose $(\pmb{t},\pmb{q})\in SE(3)$ of a camera expressed in a world reference frame, where $\pmb{t}=\!\!\,^w\pmb{t}_c \in \mathbb{R}^3$ and $\pmb{q}=\!\!\,^w\pmb{q}_c \in SO(3)$ are respectively the 3D position and quaternion vector representing the orientation of the camera in the world frame.

In all end-to-end deep learning based methods, the cornerstone is the loss function that embeds into a scalar the error between an estimated pose $(\pmb{\hat{t}},\pmb{\hat{q}})$ and the ground truth $(\pmb{t},\pmb{q})$ in $SE(3)$. We make a review of existing loss functions and present their characteristics.

\subsection{Loss functions}
\paragraph{PoseNet~\cite{kendall_posenet_2016}} this loss function weights the contribution of translation and rotation errors into a single quantity using a scale factor:
\begin{equation}\label{eq:posenetloss}
    \mathcal{L}_P = \left\Vert \pmb{\hat{t}} - \pmb{t}\right\Vert_2 + \beta \left\Vert \pmb{\hat{q}} - \frac{\pmb{q}}{\left\Vert \pmb{q} \right\Vert}\right\Vert_2
\end{equation}
where $\beta$ is a positive scalar that weights rotation importance over translation.

\paragraph{Homoscedastic uncertainty~\cite{kendall_geometric_2017,kendall_multi-task_2018}} tries to reach an optimal balance between rotation and translation errors. It is achieved by optimizing global scalars $\hat{s}_t$ and $\hat{s}_q$ through backpropagation of the following loss function:
\begin{equation}
    \mathcal{L}_{HU} = \left\Vert \pmb{\hat{t}} - \pmb{t}\right\Vert_1 e^{-\hat{s}_{t}} + \hat{s}_{t} + \left\Vert \pmb{q} - \frac{\pmb{\hat{q}}}{\left\Vert \pmb{\hat{q}} \right\Vert}\right\Vert_1 e^{-\hat{s}_{q}} + \hat{s}_{q}
\end{equation}
where $\hat{s}_t$ and $\hat{s}_q$ respectively represent the natural logarithm of the translational and rotational homoscedastic task noise variance.

\paragraph{Geometric reprojection loss~\cite{kendall_geometric_2017}} a function derived from the classical reprojection error. Its training needs some 3D points of the scene for each view. Let $\pi(\pmb{t},\pmb{q},\pmb{P})$ be the function that projects a 3D point $\pmb{P}$ into the camera view with pose $(\pmb{t}, \pmb{q})$, the Geometric reprojection loss function is defined as:
\begin{equation}
    \mathcal{L}_G = \frac{1}{\vert\mathcal{G}\vert} \sum_{\pmb{P}_i \in \mathcal{G}} \left\Vert \pi(\pmb{t},\pmb{q},\pmb{P}_i) - \pi(\pmb{\hat{t}},\pmb{\hat{q}}, \pmb{P}_i) \right\Vert_1
\end{equation}
where $\mathcal{G}$ is the subset of 3D points observed by the current view.

\paragraph{MaxError~\cite{brachmann_dsac_2018}} DSAC final pose regression relies on the following loss function that we will refer to as MaxError:
\begin{equation}
    \mathcal{L}_{ME} = max\left(\measuredangle(\pmb{q},\pmb{\hat{q}}),\,\left\Vert \pmb{\hat{t}} -\pmb{t}\right\Vert\right)
\end{equation}
where $\measuredangle(\pmb{q},\pmb{\hat{q}})$ is the measured angle in degrees between rotations in 3D space induced by $\pmb{q}$ and $\pmb{\hat{q}}$, $\pmb{t}$ and $\pmb{\hat{t}}$ are expressed in $cm$ in order to reduce the magnitude gap between translation and rotation. Note that in the original paper, this loss is only applied as a step in their full relocalization pipeline. Here, we only evaluate the performance of the loss within a much simpler end-to-end pose regressor.

\begin{figure}[t]
    \centering
    \includegraphics[width=\columnwidth]{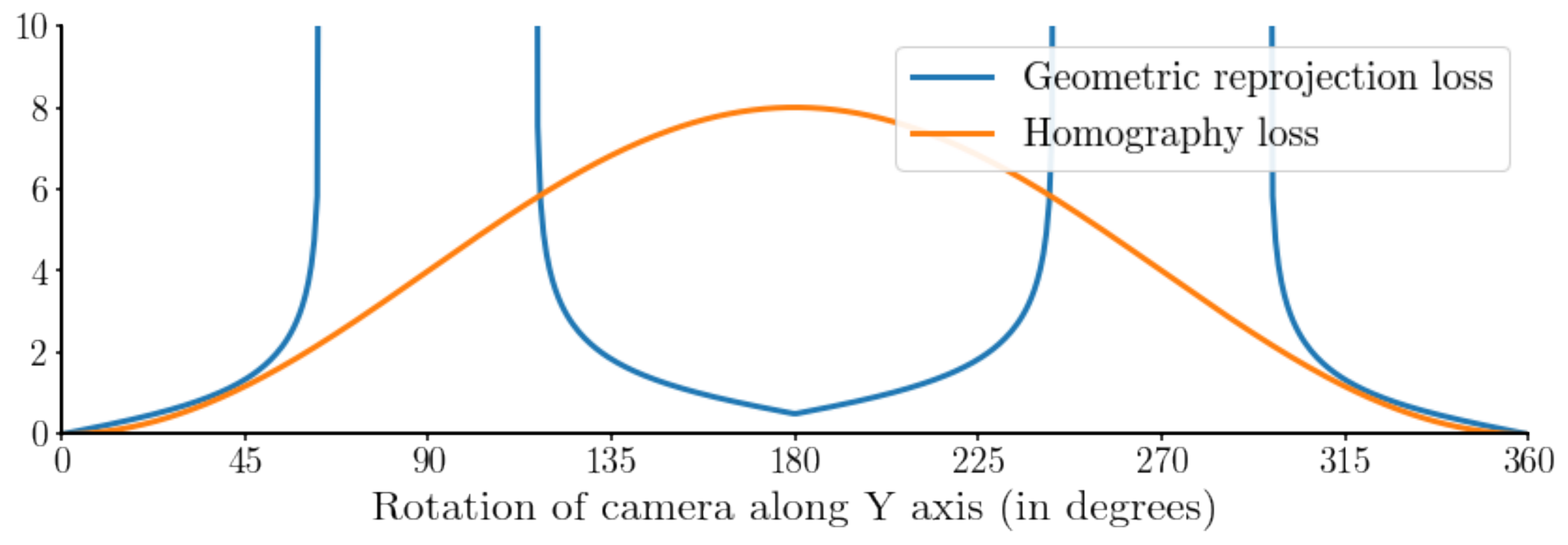}
    \caption{Evolution of the Geometric reprojection loss function (blue) and our Homographic loss (orange) when the relative pose of the ground truth and estimated cameras varies with a rotational movement along the Y axis.}
    \label{fig:geometric_error}
\end{figure}

\subsection{Loss functions characteristics}
An issue with PoseNet loss is that $\beta$ is not easily set. The rotation error is defined as the $L2$ norm between $\pmb{\hat{q}}$ and $\pmb{q}$ unit quaternions. This does not relate to any intuitive geometric phenomena. Moreover, all poses are optimized with the same relative weight between translational and rotational errors, no matter what the camera observes. Furthermore, translation and rotation errors are not comparable metrics. Fig. \ref{fig:posenet_3d} shows the influence of $\beta$ on PoseNet's loss function. A small value induces strong translational gradients but a flat profile in orientation, whereas high $\beta$ values assign importance to rotation over flat translation evolution. Given a scene, a well chosen $\beta$ allows optimizing the parameters in all dimensions. PoseNet is a muti-objective loss and presents the common problems encountered in such setting. The $\beta$ selection is not obvious and needs to be determined through trial and error. Even with a properly set $\beta$, stochastic optimization may converge to different optima on the Pareto front. 

MaxError solves the compromise between translational and rotational errors by fixing a heuristically chosen scale between them (\textit{i.e.}, translation in $cm$ and rotation in degrees). While the scale is physically interpretable, it shares the other issues with PoseNet related to the multi-objective optimization. In this case, the problem is tackled by minimizing the highest error at each step. Like PoseNet, it also shares the same global weighting between translation and rotation errors for all frames.

The Homoscedastic loss~\cite{kendall_geometric_2017, kendall_multi-task_2018} reveals characteristics similar to PoseNet and MaxError losses. However, its parameters are more robust to the initialization, since they are optimized during training.

Geometric reprojection loss implicitly solves the translation and rotation weighting problem by directly computing the reprojection error of the observed 3D points in each image. Furthermore, the contributions of rotations and translations can be found automatically and locally in each viewpoint. However, this loss is often unstable. As it will be detailed in the next section, it highly depends on the initialization of the estimated poses and on the scene points visible from the ground truth camera. Poor pose estimates and outlier points will easily lead to divergence in the case of stochastic optimization. Heuristic procedures such as pre-training initialization (using a different technique) or clipping of the error are required to stabilize training. Moreover, the loss presents a wide undesirable local minimum when points are projected on the backside of the image plane (see Fig.~\ref{fig:geometric_error}).

\begin{figure}[!b]
    \centering
    \includegraphics[width=\columnwidth]{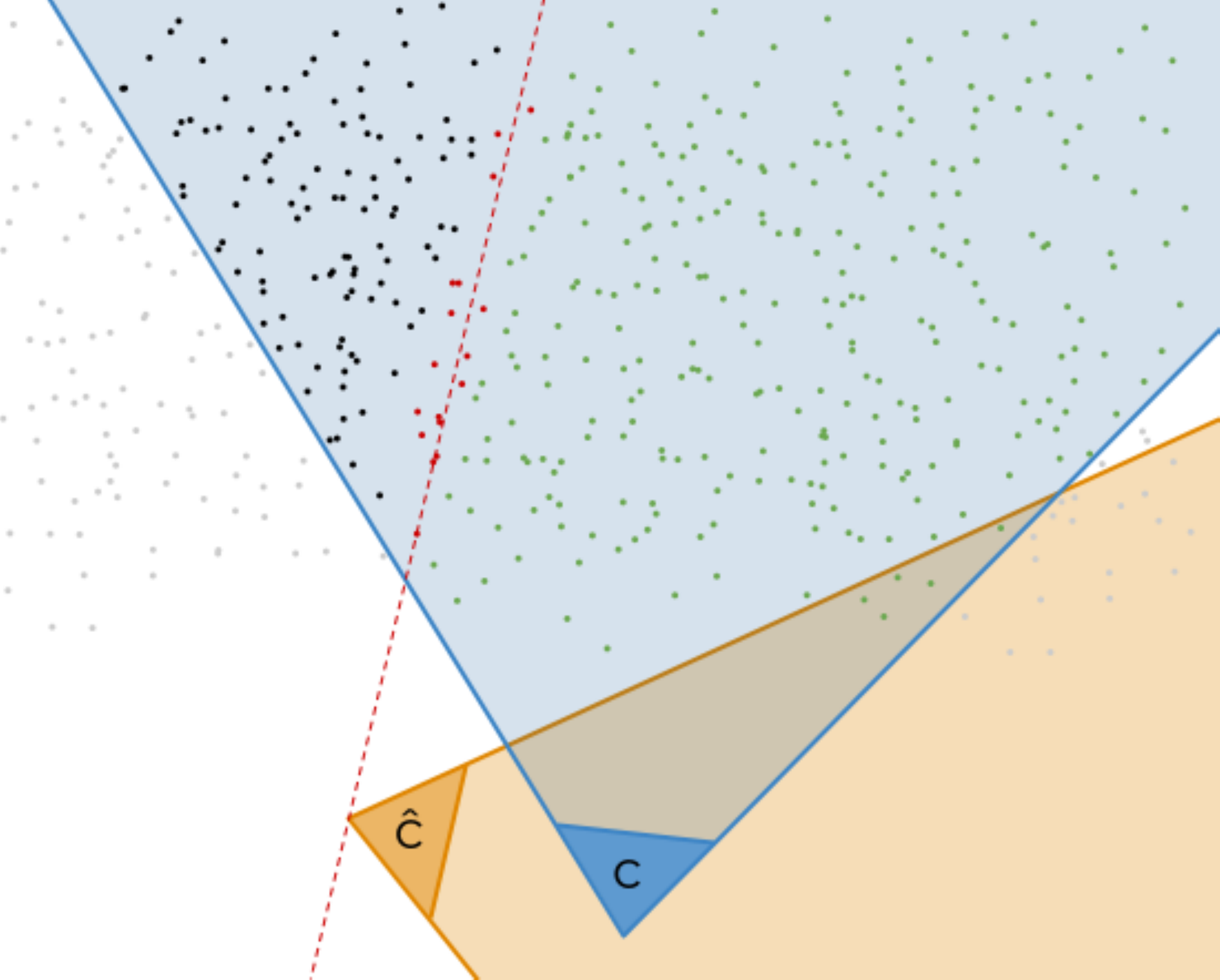}
    \caption{Top view representation of the ground truth (blue) and estimated (orange) camera views and the scene point cloud. The grey points are outside the field of view of the ground truth. The green points are in front of the image planes of the two cameras. The dashed red line represents the $x,y$ plane of the estimated camera frame. The red dots that are close to this plane are projected to infinity through the pinhole projection model. The black dots project backward from the image plane of the estimated camera.}
    \label{fig:infinite_reproj}
\end{figure}


\section{The proposed homography-based loss function}
\label{sec:method}
This section introduces a novel loss function for pose regression based on homographies between ground truth and estimated poses. Through this formulation, we aim to approximate the reprojection error while avoiding some of its drawbacks.

\subsection{Motivations for approximating the reprojection error}
The reprojection error consists in measuring the 2D distance between the projection of a set of 3D points into two camera views. If the poses are identical, then the points are superimposed. It has been widely used to solve computer vision problems such as mosaicking or 3D reconstruction \cite{hartley_multiple_2004}. Its physical meaning is easy to understand because it can be represented graphically in the image plane.
Furthermore, under the commonly used Gaussian assumption on the reprojection noise, the least square minimization of this error is equivalent to the maximum likelihood estimator which is asymptotically efficient~\cite{refregier_noise_2012}.

Its use in deep learning models is more cumbersome for multiple reasons. It relies on the choice of the 3D points that are projected on the image plane. Depending on the method used to estimate the camera ground truth poses, 3D points of the scene may already be available. However, if the scene geometry is not available, one might have to triangulate 3D points from the camera poses and 2D-2D matches. Moreover, in the initial state of the neural network, pose predictions are initialized around an arbitrary value that is often far from the ground truth. Some points can be projected to infinity if they are in the camera $(x,y)$ plane (see Fig. \ref{fig:infinite_reproj}). Reprojection error may also lead to a local minimum when 3D points are projected onto the backside of the image plane (see Fig. \ref{fig:geometric_error}). To overcome these problems, the network is usually initialized by first training it with another loss function for a few epochs~\cite{kendall_geometric_2017}. In addition, if a point is projected to infinity during the optimization, it results in an infinite loss, causing the model to diverge. In practice, this problem is often addressed by clipping reprojection error distances that exceed a threshold. However, in doing so, all clipped points lay on a flat maximum with a zero-valued gradient, therefore not contributing to the optimization.

Alternatively, the homography-based loss function we design tackles this issue by relying on the integration of homographies computed for a set of virtual parallel planes. It offers a competitive accuracy and a high numerical stability making a simple single step learning possible.

\subsection{Homography-based loss function}
\label{subsec:homographyloss}

The idea behind this work is that the 3D points used to express a pose error such as a reprojection error do not need to be real points, but can well be  a set of designated virtual (or hypothetical) points. To eliminate the infinite error problem, we could, for instance, regularly sample virtual 3D points located in front of the image planes of both cameras. Nevertheless, as the poses become more distant from each other, shared 3D observations become scarce and virtual point sampling becomes increasingly difficult.

Following this rationale, but eliminating the problems related to the choice of points, we can cut the scene into planes containing an infinity of these points. We can then calculate our error directly in the homographies induced by these planes. The homography being defined as the transformation of the same 3D plane from one projective view to another. That is, given a plane in the 3D scene and its projected points in a camera view, the homography maps these points into another camera view.

When the poses are superimposed, their homography induced by any plane not containing the center of the cameras is equal to the identity matrix. In Appendix~\ref{ap:proof}, we extend this property by demonstrating that, given a specific $^c\pmb{n}$, poses are superimposed if and only if their homography is equal to the identity matrix. Thus, for a given plane, we may quantify the error as the difference between the identity matrix and the homography induced by that plane between the ground truth and the estimated poses. We define the error for that plane as the squared Frobenius norm of the resulting difference matrix. We then integrate these errors for all possible planes between two given boundaries (see Fig. \ref{fig:homography_loss}).

Let us define the homography~\cite{hartley_multiple_2004}:
\begin{equation}
    ^{\hat{c}}\pmb{H}_c = \!\,^{\hat{c}}\pmb{R}_c - \!^{\hat{c}}\pmb{t}_c \,^c\pmb{n}^T/\,x
    \label{eq:homography}
\end{equation}
where $^c\pmb{n}$ is the normal to the considered plane expressed in the ground truth camera frame, $x$ is the distance to that plane and $^{\hat{c}}\pmb{R}_c$, $^{\hat{c}}\pmb{t}_c$ are the rotation and translation of the ground truth camera expressed in the estimated camera frame. For the sake of clarity, we will further use the notation $\pmb{H} = \,\!^{\hat{c}}\pmb{H}_c$.

Next, we show how the homographic error, defined as the squared Frobenius norm of its difference with the identity matrix, is related to the reprojection error. Let $\pmb{P}$ be a 3D point observed by two cameras. Let $\pmb{p}=(p_x, p_y, 1)^T$ and $\pmb{p'}=(p'_x, p'_y, 1)^T$ be the 2D homogeneous representations of the projection of $\pmb{P}$ in two different camera views. The reprojection error of $\pmb{p}$ is defined as:
\begin{align}
    e(\pmb{p}) &= (p_x - p'_x)^2 + (p_y - p'_y)^2\\
               &= (\pmb{p} - \pmb{p'})^T (\pmb{p} - \pmb{p'})\label{eq:reproj_error}
\end{align}
We now assume that $\pmb{P}$ is in a plane that induces a homography $\pmb{H}$ between the two camera views. We want to retrieve the reprojection error by expressing $\pmb{p'}$ in terms of $\pmb{H}\pmb{p}$. Let us explicit $\pmb{H}$ components:
\begin{equation}
    \pmb{H} = 
    \begin{pmatrix}
    h_{11} & h_{12} & h_{13}\\
    h_{21} & h_{22} & h_{23}\\
    h_{31} & h_{32} & 1
    \end{pmatrix}
\end{equation}
We note $\pmb{p''}$ the 2D homogeneous point resulting from $\pmb{H}\pmb{p}$:
\begin{equation}
    \pmb{p''} = \pmb{H}\pmb{p} = (p''_x, p''_y, s)^T
\end{equation}
where $s=h_{31} p_x + h_{32} p_y + 1$.

\noindent By the definition of the homography, $\pmb{H}\pmb{p} \sim \pmb{p'}$ in homogeneous coordinates. Thus, in the euclidean space:
\begin{equation}\label{eq:euclidean_hp}
    \pmb{p'} = \frac{\pmb{p''}}{s} = \frac{\pmb{H}\pmb{p}}{s}
\end{equation}
When replacing (\ref{eq:euclidean_hp}) into (\ref{eq:reproj_error}) we can express the reprojection error in terms of $\pmb{H}$:
\begin{align}
    e(\pmb{p}) &= \left(\pmb{p} - \frac{\pmb{H}\pmb{p}}{s}\right)^T \left(\pmb{p} - \frac{\pmb{H}\pmb{p}}{s}\right)\\
               &= \pmb{p}^T \left(\pmb{I} - \frac{\pmb{H} + \pmb{H}^T}{s} + \frac{\pmb{H}^T \pmb{H}}{s^2}\right) \pmb{p}\label{eq:reproj_H}
\end{align}
where $\pmb{I}$ is the identity matrix.

\noindent As the estimated pose tends towards the ground truth pose, $s$ tends towards $1$. We will use the approximation $s\approx 1$ to simplify (\ref{eq:reproj_H}). This way, our homographic error will tend to the reprojection error when poses are close. Then, (\ref{eq:reproj_H}) becomes:
\begin{equation}
    e(\pmb{p}) = \pmb{p}^T (\pmb{I} - \pmb{H})^T (\pmb{I} - \pmb{H})\,\pmb{p}
\end{equation}
While we have now expressed $e(\pmb{p})$ in terms of $\pmb{H}$, this error still relies on specific 2D points in the camera view. As we do not want our loss to rely on any specific 3D point, it should not rely either on their 2D projection. Working around this, we could consider that $\pmb{p}$ could be any point on the sensor. To cover all possibilities, we integrate the error over the entire sensor. To facilitate this task, we will take the trace of our error to use the trace cyclic property. Since our error is a scalar, it is equal to its trace:
\begin{equation}
    e(\pmb{p}) = \Tr \left(\pmb{p}^T (\pmb{I} - \pmb{H})^T (\pmb{I} - \pmb{H}) \,\pmb{p}\right)
\end{equation}
Then, we can use the cyclic property of the trace to isolate $\pmb{p}$:
\begin{equation}
    e(\pmb{p}) = \Tr \left(\pmb{p}\pmb{p}^T (\pmb{I} - \pmb{H})^T (\pmb{I} - \pmb{H}) \right)
\end{equation}
with
\begin{equation}
    \pmb{p}\pmb{p}^T = \begin{pmatrix}
    p_x^2 & p_x p_y & p_x\\
    p_y p_x & p_y^2 & p_y\\
    p_x & p_y & 1
    \end{pmatrix}
\end{equation}
Let $w$ and $h$ be the respective width and height of our sensor, we can integrate the error over all points of our sensor:
\begin{align}
    &\int_{-w/2}^{w/2}\int_{-h/2}^{h/2} \Tr \left(\pmb{p}\pmb{p}^T (\pmb{I} - \pmb{H})^T (\pmb{I} - \pmb{H}) \right) \dif p_x \dif p_y\\
    &= \Tr\left(\begin{pmatrix}
        \frac{hw^3}{12} & 0 & 0\\
        0 & \frac{wh^3}{12} & 0\\
        0 & 0 & wh
    \end{pmatrix} (\pmb{I} - \pmb{H})^T (\pmb{I} - \pmb{H}) \right)
\end{align}
This results in a diagonal matrix simply weighting the dimensions of the reprojection according to the size of the sensor. As we want our loss to be generic to the size of the sensor, we will simply drop this matrix.

We finally have our homographic error which, by definition, because $(\pmb{I}-\pmb{H})$ is real,  is equivalent to a Frobenius norm:
\begin{equation}\label{eq:homographicerror}
    \Tr \left((\pmb{I} - \pmb{H})^T (\pmb{I} - \pmb{H}) \right) = \left\Vert \pmb{I}-\pmb{H} \right\Vert_F^2
\end{equation}
We further extend the definition of the single plane homographic error (\ref{eq:homographicerror}) to a slab\footnote{defined as the set between two parallel planes as in \cite{boyd_vandenberghe_2004}}. We integrate the expression over the planes within a given range of distances and along a particular direction. Let $x_{min}$ and $x_{max}$ be the minimum and maximum distances of the planes containing our observations. We introduce the analytic form of our homography-based loss function:
\begin{equation}
    \mathcal{L}_H = \frac{1}{x_{max} - x_{min}}\int_{x_{min}}^{x_{max}}\left\Vert \pmb{I}-\pmb{H} \right\Vert_F^2 \dif x\label{eq:homographyloss_simplified}
\end{equation}
Note that we normalize the loss by the region of the considered scene dimension ($x_{max} - x_{min}$). This is because every frame has its own distance range of observations. By normalizing, we ensure that each frame cost is on the same scale.
We can then solve the integral by substitution of (\ref{eq:homography})
in (\ref{eq:homographyloss_simplified}) resulting in our final loss function:

\begin{equation}{
    \mathcal{L}_H =\!\Tr\left(\pmb{A}\! + \! \pmb{B} \,\frac{\ln\left(\frac{x_{max}}{x_{min}}\right)}{x_{max}-x_{min}} + \frac{\pmb{C}}{x_{min} \cdot x_{max}} \right)}
    \label{eq:homographyloss_final}
\end{equation}
where
\begin{align}
    \label{eq:A}\pmb{A} &= (\pmb{I} - \!\,^{\hat{c}}\pmb{R}_c) (\pmb{I} - \,\!^{\hat{c}}\pmb{R}_c)^T\\
    \label{eq:B}\pmb{B} &= \,^c\pmb{n} \,^{\hat{c}}\pmb{t}_c^T (\pmb{I} - \,\!^{\hat{c}}\pmb{R}_c) + \left(^c\pmb{n} \,^{\hat{c}}\pmb{t}_c^T (\pmb{I} - \,\!^{\hat{c}}\pmb{R}_c)\right)^T\\
    \label{eq:C}\pmb{C} &= \,^c\pmb{n} \,^{\hat{c}}\pmb{t}_c^T \left(^c\pmb{n} \,^{\hat{c}}\pmb{t}_c^T\right)^T
\end{align}

\subsection{Parameterization}\label{subsec:parameters}
By looking at (\ref{eq:homographyloss_final}), we can isolate the 5 parameters our loss relies on. $^{\hat{c}}\pmb{R}_c$ and $^{\hat{c}}\pmb{t}_c$ are directly computed from ground truth and estimated poses. We set $^c\pmb{n} = (0, 0, -1)^T$, so that all homographies are induced by planes parallel to the ground truth sensor, as if they faced the camera. This leaves us with 2 parameters, $x_{min}$ and $x_{max}$, representing the minimum and maximum distances of these planes to the ground truth sensor. We introduce two different ways to set these parameters, inspired by different uses of the loss, and leading to different implementations.

\begin{figure}[t]
    \centering
    \includegraphics[width=\columnwidth]{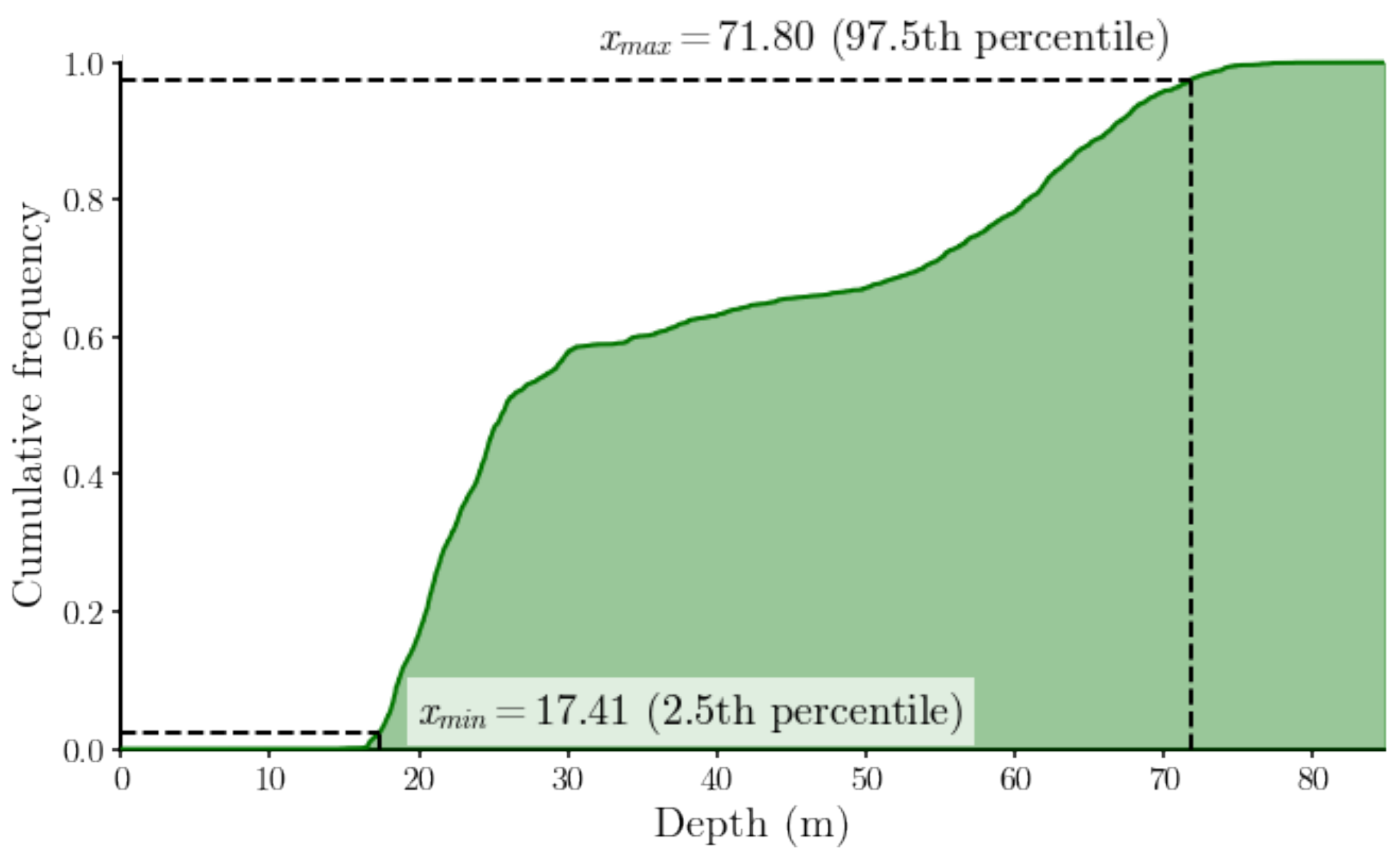}
    \caption{Cumulative histogram of the depths of the points in the scene. $x_{min}$ and $x_{max}$ depth values can be selected as the $2.5th$ and $97.5th$ percentile.}
    \label{fig:depth_histogram}
\end{figure}

The first method best approximates the reprojection error, but requires 3D data of the scene. Given this data, the two parameters can be computed for each frame. For every frame, we compute a depth histogram of its 3D observations. We then set its $x_{min}$ and $x_{max}$ parameters as a given percentile of the distribution of the depths (see Fig. \ref{fig:depth_histogram}). We refer to this method as \textit{Local homography} loss function.

The other way of setting these parameters does not require any 3D information of the scene. If no 3D data is available, it is possible to manually set global $x_{min}$ and $x_{max}$ intuitively according to the images in the dataset, by estimating minimum and maximum scene limits. Unlike the aforementioned method, these parameters will be shared by every frame. We refer to this manner of configuring the loss as \textit{Global homography} loss function. Note that if 3D data is available, it is also possible to set global $x_{min}$ and $x_{max}$ from a global depth histogram.

\section{Experiments}
\label{sec:expe}

\begin{table*}[t]
\begin{center}
\caption{Poses mean reprojection distance on train and test sets, percentage of poses localized within a given threshold}
\label{tab:results}
\resizebox{\linewidth}{!}{%
\begin{tabular}{|l|c|c|c|c|c|c|}
    \hline
    \multirow{2}{*}{\vspace{-0.35cm}Scene}&\multicolumn{4}{c|}{Existing} & \multicolumn{2}{c|}{Ours}\\\cline{2-7}
    & \makecell{PoseNet \\ ($\beta=500$)} & Homoscedastic & MaxError & \makecell{Geometric \\ reprojection loss} & \makecell{Global \\ homography} & \makecell{Local \\ homography}\\
    \hline
    Great Court & 10.5px, \redbold{117.7px}, \redbold{13\%}, \redbold{36.4\%} & \redbold{9.469px}, 148.1px, 7.6\%, 26.6\% & 220.9px, 623.9px, 0.4\%, 2.8\% & 67.46px, 182.9px, 1.1\%, 8.4\% & 92.9px, 235px, 0.7\%, 6.2\% & 143px, 261.3px, 0.4\%, 1.1\%\\
    King's College & 7.18px, 33.93px, 64.4\%, 92.7\% & 5.18px, 24.67px, 60.1\%, 92.1\% & 215.9px, 204.3px, 6.1\%, 26.8\% & 4.53px, \redbold{16.16px}, \redbold{71.7\%}, \redbold{94.2\%} & 4.8px, 23.2px, 61.2\%, 92.7\% & \redbold{4.52px}, 23.07px, 61.5\%, 91.8\%\\
    Old Hospital & 9.02px, 97.5px, 23.6\%, 56\% & 8.19px, 80.63px, 18.7\%, 56.6\% & 98.23px, 177.2px, 9.3\%, 34.6\% & 6.56px, \redbold{63.84px}, \redbold{28.6\%}, \redbold{71.4\%} & 6.76px, 100.2px, 15.9\%, 51.6\% & \redbold{6px}, 92.8px, 23.1\%, 61\%\\
    Shop Façade & 17.19px, 135.4px, 15.5\%, \redbold{68\%} & 12.18px, 125px, 14.6\%, 49.5\% & 194.8px, 218.9px, 3.9\%, 31.1\% & 10.91px, \redbold{117.4px}, \redbold{18.4\%}, 56.3\% & 10.82px, 149px, 11.7\%, 47.6\% & \redbold{10.27px}, 130.6px, 12.6\%, 58.3\%\\
    St Mary's Church & 16.34px, 161.5px, 13.4\%, 50.8\% & 13.08px, 125px, 16.8\%, 56.6\% & 109.1px, 260.1px, 2.6\%, 20.6\% & 33.51px, \redbold{104.6px}, 13.6\%, 52.6\% & 10.94px, 114.7px, 17.0\%, 56.2\% & \redbold{10.15px}, 108.5px, \redbold{18.1\%}, \redbold{57.5\%}\\
    Street & 36.29px, 790.2px, 0.4\%, 2.1\% & 32.04px, 758.2px, 0.2\%, 1.8\% & 224.4px, 768.5px, 0.1\%, 0.4\% & 390.5px, \redbold{505.2px}, 0\%, 1.1\% & 25.02px, 733.6px, 0.5\%, 2.4\% & \redbold{23.32px}, 683px, \redbold{0.6\%}, \redbold{3.5\%}\\
    \hline
    chess & 4.36px, 40.49px, 78.4\%, 91.5\% & 1.92px, 30.73px, 80.4\%, 94\% & 2.43px, 34.76px, 80.8\%, 96.4\% & 1.51px, \redbold{26.46px}, 80.9\%, 95.1\% & 1.94px, 30.27px, 82\%, \redbold{96.5\%} & \redbold{1.5px}, 28.85px, \redbold{82.3\%}, 96.2\%\\
    fire & 4.59px, 80.03px, 34.8\%, 61.3\% & 2.09px, 88.35px, 32.5\%, 66\% & 2.23px, 86.92px, \redbold{35.4\%}, \redbold{68.8\%} & 1.67px, 89.16px, 30.5\%, 65.1\% & 2px, 83.89px, 32.3\%, 66.3\% & \redbold{1.59px}, \redbold{79.05px}, 31.8\%, 64.3\%\\
    heads & 7.56px, 96.01px, 32.1\%, 57\% & 3.34px, 90.13px, 30.8\%, 53\% & 3.53px, 71.75px, 31.6\%, \redbold{62.4\%} & 2.46px, 75.13px, 31.7\%, 55\% & 2.93px, 76.22px, 33.3\%, 59.2\% & \redbold{2.34px}, \redbold{69.85px}, \redbold{33.9\%}, 58.3\%\\
    office & 4.97px, 55.03px, 60\%, 90.1\% & 2.65px, 59.25px, 54.7\%, 86.6\% & 3.06px, 59.68px, \redbold{62.3\%}, 87.6\% & 2.22px, 50.3px, 59.5\%, 90.3\% & 2.31px, 55.2px, 57.3\%, \redbold{90.7\%} & \redbold{2.05px}, \redbold{46px}, \redbold{62.3\%}, 86.1\%\\
    pumpkin & 3.38px, 121.1px, \redbold{59\%}, 74.4\% & 1.66px, 80.3px, 51.7\%, 73.2\% & 2px, 89.6px, 51.9\%, \redbold{77.2\%} & 1.35px, 87.5px, 50.9\%, 71.9\% & 1.63px, 97.3px, 50.1\%, 71.5\% & \redbold{1.2px}, \redbold{69.1px}, 53.5\%, 73.9\%\\
    redkitchen & 4.59px, 70.71px, 45.1\%, 74.6\% & 2.24px, 89.24px, 45.8\%, 73.8\% & 3px, 79.38px, 54.9\%, 79.8\% & \redbold{1.9px}, 83.71px, 48.4\%, 77\% & 2.85px, 78.69px, 50.5\%, 75.4\% & 2.11px, \redbold{66.31px}, \redbold{57.1\%}, \redbold{81.7\%}\\
    stairs & 4.24px, 123.4px, 12.7\%, 36.6\% & 1.87px, 127px, 13.5\%, 58.6\% & 2.26px, 133.1px, \redbold{22.2\%}, \redbold{59\%} & 1.65px, 153.9px, 4\%, 28.1\% & 1.75px, 144.9px, 18.4\%, 57.9\% & \redbold{1.49px}, \redbold{120.9px}, 17\%, 55.8\%\\
    \hline
\end{tabular}}
\end{center}
\end{table*}

\begin{table}[t]
\begin{center}
\caption{Comparing our implementation with~\cite{kendall_geometric_2017} on King's College}
\label{tab:comparison}
\begin{tabular}{|l|c|c|c|c|c|}
    \hline
    & \makecell{PoseNet\\$\beta=500$} & Homoscedastic & Geometric\\
    \hline
    Kendall \textit{et al.}~\cite{kendall_geometric_2017} & 1.66m, 4.86° & 0.99m, 1.06° & 0.88m, 1.04°\\
    Ours & 0.76m, 0.90° & 0.87m, 1.15° & 0.64m, 0.89° \\
    \hline
\end{tabular}
\end{center}
\end{table}

We implemented PoseNet~\cite{kendall_posenet_2016}, Homoscedastic~\cite{kendall_geometric_2017, kendall_multi-task_2018}, Geometric~\cite{kendall_geometric_2017} and MaxError~\cite{brachmann_dsac_2018} losses. We tested all losses on Cambridge~\cite{kendall_posenet_2016} and 7-Scenes~\cite{shotton2013_7scenes} datasets with a MobileNetV2~\cite{sandler_mobilenetv2_2019} architecture. We compared the results between them and with previously published ones.

\subsection{Datasets}
In~\cite{brachmann2021limits}, Brachmann \textit{et al.} show that the performance of a relocalization method on a given dataset is greatly impacted by the method used to build the ``ground truth" of this dataset. Very popular methods for estimating the camera poses, and the scene geometry are Structure-from-Motion (SfM) and depth-based SLAM. In our case, some losses might be advantaged by one method or the other. To try to alleviate this issue when benchmarking the different losses, we evaluated them on two datasets whose ground truth poses were estimated using different methods. Cambridge dataset was built using SfM and consists of 6 outdoor scenes of Cambridge city while 7-Scenes was built using depth-based SLAM and provides 7 different indoor scenes. All scenes in both datasets are visited several times, and train and test sequences consist of different visits.

\subsection{Network}
Kendall \textit{et al.} used GoogLeNet~\cite{szegedy_going_2014} as a regression backbone. They then replaced the classification head with 2 dense layers with respective feature sizes of 2048 and 7 (3 for translation and 4 for the quaternion). In this paper, we use the MobileNetV2~\cite{sandler_mobilenetv2_2019} backbone provided by PyTorch and proceed to the same replacement. We chose the MobileNetV2 backbone for its versatility. Similarly to Kendall \textit{et al.}, we load weights from MobileNetV2 pre-trained on ImageNet which are available from the PyTorch Hub. Note that we normalize our input images, as suggested by PyTorch. However, we do not crop our resized image to stay consistent with~\cite{kendall_geometric_2017}. In PoseNet~\cite{kendall_posenet_2016}, the network is reportedly trained on random crops of the resized images. We tested this approach and found that it greatly deteriorates the results. We suggest that by applying a random crop to the image, it artificially moves the optical center of the camera, which might not be ideal when trying to estimate its pose. We deliberately chose to test our loss function on a simple and unique end-to-end network rather than on a more complete pipeline like DSAC~\cite{brachmann_dsac_2018} or DSAC++~\cite{brachmann_learning_2018}. Our aim is to provide an alternative to the existing pose regression loss functions, not to provide a complete pose regression system. Therefore, this study compares the loss functions on a single regression model, to easily compare and reproduce results. The use of this loss in the full pipeline of existing state-of-the-art relocalization systems remains in the scope of future work.

\subsection{Losses specifications}
\label{subsec:losses_spec}
We train the model in a common mini-batch setting, where each optimization step is based on the average loss over all camera poses of the batch. For the PoseNet loss, we fix \mbox{$\beta=500$} to compare our results with those reported in~\cite{kendall_geometric_2017}. For the Homoscedastic loss, we initialize its parameters as suggested by Kendall \textit{et al.}, $\hat{s}_t=0.0$ and $\hat{s}_q=-3.0$. As discussed earlier, when implementing the Geometric reprojection loss, we clip the reprojection error of each point at 100 to avoid divergence. As for MaxError loss, estimated quaternions always ended up converging towards the null vector on the Cambridge dataset. We fixed this by adding a regularization term, $MSE\left(\Vert q \Vert_2, 1\right)$, to the final loss. Nevertheless, we only reached $\sim 1.2k-3.5k$ epochs before the loss diverges. For our homographic losses, we chose $x_{min}$ and $x_{max}$ as a percentile of the depth distribution of 3D points in the scene, as discussed in section (\ref{subsec:parameters}). $x_{min}$ and $x_{max}$ were set to the $2.5th$ and $97.5th$ percentiles, respectively.

\subsection{Training}
We train all models with an Adam optimizer~\cite{kingma_adam_2017, loshchilov_decoupled_2019} with a learning rate of $10^{-4}$. We train for 5k epochs with a batch size of 64, dropping the last batch if smaller. During our experiments, we found that for the homography losses, an Adam's epsilon of $10^{-14}$ instead of the default $10^{-8}$ leads to better results. This is because at the end of the optimization, our losses reach very low values $\sim 10^{-4}$. For the Geometric reprojection loss, we initialize the network by training for 500 epochs on the Homoscedastic loss.

\section{Results and discussion}
\label{sec:discussion}

In Table \ref{tab:comparison} we report median translation and rotation errors on different losses and compare them with previously published ones. We show that our implementation's results are consistent with PoseNet~\cite{kendall_posenet_2016,kendall_geometric_2017}. Small differences may be due to the use of MobileNetV2 instead of GoogLeNet as a backbone. PoseNet differences can be explained by the cropping done in the original implementation \cite{kendall_posenet_2016} which is no longer performed in more recent studies~\cite{kendall_geometric_2017}.

Table \ref{tab:results} presents the results we obtained on Cambridge and 7-Scenes datasets. We report two different types of metrics. The first criterion is the mean reprojection distance. For a given camera, we select the 3D points that it observes as given by our ground truth. We then compute the L2 norm between the projection of these points on the ground truth and the estimated camera view. We clip the reprojection distance of each point at $1000$px to reduce the impact of outliers on the metric. Finally, we compute the mean of all these distances. The second criterion is the percentage of poses localized within a given threshold in meters and in degrees. For each loss function and for each scene, we report i) the mean reprojection distance on the train set, ii) the mean reprojection distance on the test set, iii) \& iv) the second criterion with different threshold values on the test set. We use different threshold values for Cambridge and 7-Scenes datasets because the ratio between average translation and rotation errors is not the same. For Cambridge, thresholds are 2m/2° and 3m/5°. For 7-Scenes, thresholds are 0.25m/10° and 0.5m/15°.

We argue that the different metrics may be more favorable to different losses. When evaluating poses independently of what the camera observes, the percentage of poses localized within a given threshold might be a better indicator of performance than the mean reprojection distance. On the other hand, when evaluating poses based on their shared observations, the mean reprojection distance is a better indicator of whether images captured from estimated and ground truth cameras poses result in the same view. Both evaluation methods are more or less important depending on the application, \textit{e.g.}, for ego-motion we might need to perform best on the percentage criterion, while in augmented reality we may seek to improve the mean reprojection distance. This is why, in addition to the percentage of poses localized within a given threshold classically reported in previous work~\cite{sarlin_coarse_2019,sarlin_back_2021,sattler2018benchmarking}, we also report mean square reprojection distance on the train and test sets. We choose to report this error on the train dataset because this way we directly monitor how the losses optimize it.

Overall, we notice that the Geometric reprojection loss performs better on the Cambridge dataset, while our homography loss shows the best results on the 7-Scenes dataset. As argued by Brachmann \textit{et al.}~\cite{brachmann2021limits}, this could be explained by the relation between the method used to estimate the ``ground truth'' poses and the cost minimized by the loss function. As Cambridge ground truth poses were obtained using SfM, the Geometric reprojection loss minimizes the same quantities using the exact same data as the ones used to estimate the ground truth. Conversely, 7-Scenes poses were obtained using depth-based SLAM. While our loss does not minimize the same quantities as depth-based SLAM, it might greatly benefit from the access to dense depth maps since its parameters can be directly deduced from them. Interestingly, the proposed homography-based loss shows the best results with regard to the reprojection distance on the training set, which is by definition what the Geometric reprojection loss minimizes. This could be explained by the improved stability and convexity of our proposed losses in a mini-batch stochastic gradient descent training setting.

\section{Conclusion}
\label{sec:conclusion}
We introduced a new loss function for camera pose regression, which is based on the integration of homographies’ virtual planes between the minimum and maximum scene distances to the sensor. It relies on two physically interpretable parameters that can either be tuned manually or computed from 3D data. Moreover, it requires no prior initialization to converge. The obtained results show that it optimizes best the mean reprojection distance on the train set than any other loss. Depending on the application, it provides a competitive drop-in alternative to existing pose regression losses. Our loss might also be a good replacement to the Geometric reprojection loss if 3D data is not available or if the target application needs to regress a pose with regard to the camera observations without relying on specific 3D points, \textit{e.g.}, learning features. Future work will concentrate in testing the proposed loss in more complete relocalization pipelines like DSAC~\cite{brachmann_dsac_2018}, DSAC++~\cite{brachmann_learning_2018} or PixLoc~\cite{sarlin_back_2021}.


%

\appendices
\section{}\label{ap:proof}
For ease of reading, we will use the following notations in the appendix: $\pmb{t} = \,\!\! ^{\hat{c}}\pmb{t}_c$, $\pmb{R} = \,\!\!^{\hat{c}}\pmb{R}_c$, $\pmb{H} =\,\!\!^{\hat{c}}\pmb{H}_c$ and $\pmb{n} = \,\!\!^c\pmb{n}$.

In this appendix, we demonstrate that our loss function only reaches its minimum when poses are superimposed. That is, when $\pmb{R}=\pmb{I}$ and $\pmb{t}=\pmb{0}_3$. Note that our loss considers planes parallel to our sensor image plane. Therefore, our proof is only valid for $n = (0, 0, -1)^T$.

Given any $3 \times 3$ matrix $\pmb{M} \in \mathcal{R}^{3 \times 3}$, its squared Frobenius norm is
\begin{equation}
    \left\Vert \pmb{M} \right\Vert_F^2 = \sum_{i=1}^{3}\sum_{j=1}^{3} m_{ij}^2\label{eq:fro_norm}
\end{equation}
where $m_{ij}$ is the $i$th row, $j$th column element of $\pmb{M}$. Because the squared Frobenius norm of such a matrix is the sum of the square of all its elements, it is clear that $\Vert\pmb{I} - \pmb{H}\Vert_F^2$ can only be positive. It is also clear that its integration over a positive interval can only be positive and therefore, $\mathcal{L}_H \geq 0$ (\ref{eq:homographyloss_simplified}). We also know that the homography between two superimposed views is the identity matrix
\begin{equation}
[\pmb{R}, \pmb{t}] = [\pmb{I}, \pmb{0}_3] \Rightarrow \pmb{H} = \pmb{I} \Rightarrow \mathcal{L}_H = 0\label{eq:global_imply}
\end{equation}
Because $\mathcal{L}_H \geq 0$, that means that its minimum is $0$.

From (\ref{eq:fro_norm}), we can deduce that
\begin{equation}
    \mathcal{L}_H = 0 \Leftrightarrow \pmb{H} = \pmb{I}\label{eq:loss_I_H}
\end{equation}
which means that our loss reaches its minimum only when the homography is the identity matrix. It remains to prove that $\pmb{H} = \pmb{I} \Rightarrow [\pmb{R}, \pmb{t}] = [\pmb{I}, \pmb{0}_3]$.

By injecting (\ref{eq:homography}) in (\ref{eq:loss_I_H})
\begin{equation}
    \pmb{R} = \pmb{I} + \frac{\pmb{tn}^T}{d}\label{eq:R}
\end{equation}

\noindent And because $\pmb{R} \in SO(3)$ it has the property $\pmb{RR}^T = \pmb{I}$. We can use this property to constrain $\pmb{t}$.
\begin{align}
    \left(\pmb{I} + \frac{\pmb{tn}^T}{d}\right)\left(\pmb{I} + \frac{\pmb{tn}^T}{d}\right)^T &= \pmb{I}\\
    \Leftrightarrow \pmb{tn}^T + \pmb{nt}^T + \frac{\pmb{tn}^T\pmb{nt}^T}{d} &= \pmb{0}_{3 \times 3}\label{eq:so3id}
\end{align}
We fix $\pmb{n} = (0, 0, -1)^T$ and note $\pmb{t} = (t_x, t_y, t_z)^T$. We can decompose (\ref{eq:so3id})
\begin{align}
    \pmb{tn}^T &=\begin{pmatrix}
    0 & 0 & -t_x\\
    0 & 0 & -t_y\\
    0 & 0 & -t_z
    \end{pmatrix}\label{eq:tn}\\
    \pmb{nt}^T &= \begin{pmatrix}
    0 & 0 & 0\\
    0 & 0 & 0\\
    -t_x & -t_y & -t_z
    \end{pmatrix}\label{eq:nt}\\
    \pmb{tn}^T\pmb{nt}^T &= \begin{pmatrix}
    t_x^2 & t_x t_y & t_x t_z\\
    t_y t_x & t_y^2 & t_y t_z\\
    t_z t_x & t_z t_y & t_z^2
    \end{pmatrix}\label{eq:tnnt}
\end{align}
From (\ref{eq:so3id}), (\ref{eq:tn}), (\ref{eq:nt}) and (\ref{eq:tnnt}) we deduce
\begin{align}
    t_x &= 0\\
    t_y &= 0
\end{align}
And two possibilities for $t_z$
\begin{equation}
    \left\{ \begin{array}{ll}
    t_z = 0\\
    t_z = 2d
    \end{array}\right.
\end{equation}
We can further constrain $\pmb{t}$ by using another property of $SO(3)$, that is, $\det (\pmb{R}) = 1$. From (\ref{eq:R}) and (\ref{eq:tn}) with $t_x=0$, $t_y=0$ and $d \neq 0$
\begin{align}
    &\det (\pmb{R}) = \det \left(\begin{pmatrix}
    1 & 0 & 0\\
    0 & 1 & 0\\
    0 & 0 & 1 - \frac{t_z}{d}
    \end{pmatrix}\right) = 1 - \frac{t_z}{d}\\
    &\Leftrightarrow 1 - \frac{t_z}{d} = 1 \Leftrightarrow t_z = 0
\end{align}
We have just shown that $\pmb{H} = \pmb{I} \Rightarrow \pmb{t} = \pmb{0}_3$. From (\ref{eq:R}) we find that
\begin{equation}
    \pmb{R} = \pmb{I} + \frac{\pmb{0}_3\pmb{n}^T}{d} = \pmb{I}
\end{equation}
Which means that
\begin{equation}
    \pmb{H} = \pmb{I} \Rightarrow [\pmb{R}, \pmb{t}] = [\pmb{I}, \pmb{0}_3] \label{eq:HIimpliesRt}
\end{equation}
By putting (\ref{eq:global_imply}), (\ref{eq:loss_I_H}) and (\ref{eq:HIimpliesRt}) together we find
\begin{equation}
    \mathcal{L}_H = 0 \Leftrightarrow [\pmb{R}, \pmb{t}] = [\pmb{I}, \pmb{0}_3]
\end{equation}
Our loss minimum is only reached when poses are superimposed.

\ifCLASSOPTIONcaptionsoff
  \newpage
\fi



\newpage
\bibliographystyle{IEEEtran}
\bibliography{IEEEabrv,cameraready.bbl}

\begin{thebibliography}{10}
\providecommand{\url}[1]{#1}
\csname url@rmstyle\endcsname
\providecommand{\newblock}{\relax}
\providecommand{\bibinfo}[2]{#2}
\providecommand\BIBentrySTDinterwordspacing{\spaceskip=0pt\relax}
\providecommand\BIBentryALTinterwordstretchfactor{4}
\providecommand\BIBentryALTinterwordspacing{\spaceskip=\fontdimen2\font plus
\BIBentryALTinterwordstretchfactor\fontdimen3\font minus
  \fontdimen4\font\relax}
\providecommand\BIBforeignlanguage[2]{{%
\expandafter\ifx\csname l@#1\endcsname\relax
\typeout{** WARNING: IEEEtran.bst: No hyphenation pattern has been}%
\typeout{** loaded for the language `#1'. Using the pattern for}%
\typeout{** the default language instead.}%
\else
\language=\csname l@#1\endcsname
\fi
#2}}

\bibitem{piasco_survey_2018}
N.~Piasco, D.~Sidibé, C.~Demonceaux, and V.~Gouet-Brunet, ``A survey on
  visual-based localization: On the benefit of heterogeneous data,''
  \emph{Pattern Recognition}, vol.~74, pp. 90--109, 2018.

\bibitem{mur-artal_orb-slam_2015}
R.~Mur-Artal, J.~M.~M. Montiel, and J.~D. Tardós, ``Orb-slam: A versatile and
  accurate monocular slam system,'' \emph{IEEE Transactions on Robotics},
  vol.~31, no.~5, pp. 1147--1163, 2015.

\bibitem{sattler_efficient_2017}
T.~Sattler, B.~Leibe, and L.~Kobbelt, ``Efficient \& effective prioritized
  matching for large-scale image-based localization,'' \emph{IEEE Transactions
  on Pattern Analysis and Machine Intelligence}, vol.~39, no.~9, pp.
  1744--1756, 2017.

\bibitem{szegedy_going_2014}
C.~Szegedy, W.~Liu, Y.~Jia, P.~Sermanet, S.~Reed, D.~Anguelov, D.~Erhan,
  V.~Vanhoucke, and A.~Rabinovich, ``Going deeper with convolutions,'' in
  \emph{Proceedings of the IEEE Conference on Computer Vision and Pattern
  Recognition (CVPR)}, June 2015.

\bibitem{sandler_mobilenetv2_2019}
M.~Sandler, A.~Howard, M.~Zhu, A.~Zhmoginov, and L.-C. Chen, ``Mobilenetv2:
  Inverted residuals and linear bottlenecks,'' in \emph{Proceedings of the IEEE
  Conference on Computer Vision and Pattern Recognition (CVPR)}, June 2018.

\bibitem{kendall_posenet_2016}
A.~Kendall, M.~Grimes, and R.~Cipolla, ``Posenet: A convolutional network for
  real-time 6-dof camera relocalization,'' in \emph{Proceedings of the IEEE
  International Conference on Computer Vision (ICCV)}, December 2015.

\bibitem{kendall_geometric_2017}
A.~Kendall and R.~Cipolla, ``Geometric loss functions for camera pose
  regression with deep learning,'' in \emph{Proceedings of the IEEE Conference
  on Computer Vision and Pattern Recognition (CVPR)}, July 2017.

\bibitem{brachmann_dsac_2018}
E.~Brachmann, A.~Krull, S.~Nowozin, J.~Shotton, F.~Michel, S.~Gumhold, and
  C.~Rother, ``Dsac - differentiable ransac for camera localization,'' in
  \emph{Proceedings of the IEEE Conference on Computer Vision and Pattern
  Recognition (CVPR)}, July 2017.

\bibitem{brachmann_learning_2018}
E.~Brachmann and C.~Rother, ``Learning less is more - 6d camera localization
  via 3d surface regression,'' in \emph{Proceedings of the IEEE Conference on
  Computer Vision and Pattern Recognition (CVPR)}, June 2018.

\bibitem{sarlin_coarse_2019}
P.-E. Sarlin, C.~Cadena, R.~Siegwart, and M.~Dymczyk, ``From coarse to fine:
  Robust hierarchical localization at large scale,'' in \emph{Proceedings of
  the IEEE/CVF Conference on Computer Vision and Pattern Recognition (CVPR)},
  June 2019.

\bibitem{sarlin_back_2021}
P.-E. Sarlin, A.~Unagar, M.~Larsson, H.~Germain, C.~Toft, V.~Larsson,
  M.~Pollefeys, V.~Lepetit, L.~Hammarstrand, F.~Kahl, and T.~Sattler, ``Back to
  the feature: Learning robust camera localization from pixels to pose,'' in
  \emph{Proceedings of the IEEE/CVF Conference on Computer Vision and Pattern
  Recognition (CVPR)}, June 2021, pp. 3247--3257.

\bibitem{kendall_multi-task_2018}
A.~Kendall, Y.~Gal, and R.~Cipolla, ``Multi-task learning using uncertainty to
  weigh losses for scene geometry and semantics,'' in \emph{Proceedings of the
  IEEE Conference on Computer Vision and Pattern Recognition (CVPR)}, June
  2018.

\bibitem{shotton2013_7scenes}
J.~Shotton, B.~Glocker, C.~Zach, S.~Izadi, A.~Criminisi, and A.~Fitzgibbon,
  ``Scene coordinate regression forests for camera relocalization in rgb-d
  images,'' in \emph{Proc. Computer Vision and Pattern Recognition
  (CVPR)}.\hskip 1em plus 0.5em minus 0.4em\relax IEEE, June 2013.

\bibitem{hartley_multiple_2004}
R.~Hartley and A.~Zisserman, \emph{Multiple View Geometry in Computer Vision},
  2nd~ed.\hskip 1em plus 0.5em minus 0.4em\relax Cambridge University Press,
  2004.

\bibitem{refregier_noise_2012}
P.~R{\'e}fr{\'e}gier, \emph{Noise theory and application to physics: from
  fluctuations to information}.\hskip 1em plus 0.5em minus 0.4em\relax Springer
  Science \& Business Media, 2004.

\bibitem{boyd_vandenberghe_2004}
S.~Boyd and L.~Vandenberghe, \emph{Convex Optimization}.\hskip 1em plus 0.5em
  minus 0.4em\relax Cambridge University Press, 2004.

\bibitem{brachmann2021limits}
E.~Brachmann, M.~Humenberger, C.~Rother, and T.~Sattler, ``{On the Limits of
  Pseudo Ground Truth in Visual Camera Re-Localization},'' in
  \emph{International Conference on Computer Vision (ICCV)}, 2021.

\bibitem{kingma_adam_2017}
D.~P. Kingma and J.~Ba, ``Adam: {A} method for stochastic optimization,'' in
  \emph{3rd International Conference on Learning Representations, {ICLR} 2015,
  San Diego, CA, USA, May 7-9, 2015, Conference Track Proceedings}, Y.~Bengio
  and Y.~LeCun, Eds., 2015.

\bibitem{loshchilov_decoupled_2019}
I.~Loshchilov and F.~Hutter, ``Decoupled weight decay regularization,'' in
  \emph{7th International Conference on Learning Representations, {ICLR} 2019,
  New Orleans, LA, USA, May 6-9, 2019}, 2019.

\bibitem{sattler2018benchmarking}
T.~Sattler, W.~Maddern, C.~Toft, A.~Torii, L.~Hammarstrand, E.~Stenborg,
  D.~Safari, M.~Okutomi, M.~Pollefeys, J.~Sivic, F.~Kahl, and T.~Pajdla,
  ``Benchmarking 6dof outdoor visual localization in changing conditions,'' in
  \emph{Proceedings of the IEEE Conference on Computer Vision and Pattern
  Recognition (CVPR)}, June 2018.

\end{thebibliography}
%




\end{document}